\RequirePackage{fix-cm}
\documentclass[smallextended]{svjour3}       % onecolumn (second format)
\smartqed  % flush right qed marks, e.g. at end of proof
\usepackage{graphicx}
\usepackage{amsmath}
\usepackage{url}
\usepackage{booktabs}
\usepackage{multirow}
\begin{document}

\title{ATR4S: Toolkit with State-of-the-art Automatic Terms Recognition Methods in Scala
\thanks{The reported study was supported by RFBR, research project No. 14-07-00692}
}
%\subtitle{Do you have a subtitle?\\ If so, write it here}

\titlerunning{ATR4S: Toolkit with State-of-the-art ATR Methods in Scala}        % if too long for running head

\author{Nikita Astrakhantsev        
}

%\authorrunning{Short form of author list} % if too long for running head

\institute{N. Astrakhantsev \at
              ul. Solzhenitsyna 25, Moscow, 109004 Russia \\
              Tel.: +7(495)912-56-59 (ext. 461)\\
              Fax: +7(495)912-15-24\\
              \email{astrakhantsev@ispras.ru}           %  \\
%             \emph{Present address:} of F. Author  %  if needed
}

\date{Received: date / Accepted: date}
% The correct dates will be entered by the editor

\maketitle

\begin{abstract}
Automatically recognized terminology is widely used for various domain-specific texts processing tasks, such as machine translation, information retrieval or ontology construction.
However, there is still no agreement on which methods are best suited for particular settings and, moreover, there is no reliable comparison of already developed methods.
We believe that one of the main reasons is the lack of state-of-the-art methods implementations, which are usually non-trivial to recreate.

%As a result, researches tend to call old methods developed in 90s as state-of-the-art.

%it is hard to say which methods are state-of-the-art, because of no such comparison

In order to address these issues, we present ATR4S, an open-source software written in Scala that comprises more than 15 methods for automatic terminology recognition (ATR) and implements the whole pipeline from text document preprocessing, to term candidates collection, term candidates scoring, and finally, term candidates ranking.
It is highly scalable, modular and configurable tool with support of automatic caching.

We also compare 13 state-of-the-art methods on 7 open datasets by average precision and processing time.
Experimental comparison reveals that no single method demonstrates best average precision for all datasets and that other available tools for ATR do not contain the best methods.

% \PACS{PACS code1 \and PACS code2 \and more}
% \subclass{MSC code1 \and MSC code2 \and more}
\end{abstract}

\section{Introduction}
\label{intro}

Automatic terminology recognition (ATR) aims at extraction of terms --- words and collocations designating domain-specific concepts --- from a collection of text documents belonging to that domain.
Extracted terms then can be used for many tasks including machine translation~\cite{gaussier1998flow}, information retrieval~\cite{lingpeng2005improving}, sentiment analysis~\cite{mayorov2015aspect}, ontology construction and ontology enrichment~\cite{astrakhantsev2015methods}.

Despite this importance, the ATR task is still far from being solved: 
researches continue to propose new methods for ATR, which usually show average precision below 80\% even on top 500-1000 terms~\cite{syrcodis2013,zhang2008,zhang2016jate} and thus are hardly used in practice.  
%Some of the reasons are inherent to ATR due to fuzziness in term definition and its domain- and application-specificity. However, another problem 
%one of the reason - domain-specificity and application-specificity
Moreover, there is still no fair and reliable comparison of already developed methods. 
Most works compare 1-2 newly proposed methods with several old baselines on 1-3 datasets only.

The reason is that a lot efforts are required both to obtain datasets for comparison and to reimplement modern methods, which are usually non-trivial and dependent on proprietary modules.

To address these issues, we present an open-source implementation of ATR methods and their comparison on many datasets.
In more detail, our main contributions are the following:
\begin{enumerate}
\item ATR4S\footnote{\url{https://github.com/ispras/atr4s}}: an open-source implementation of more than 10 state-of-the-art methods for ATR written in Scala; 
note that it can be easily used in any language working on Java Virtual Machine including Java itself. % with Apache Spark
%- capable of extracting term of any length

\item Modification of KeyConceptRelatedness method~\cite{astrakhantsev2015phd} by replacing proprietary module for semantic relatedness computation with open-source tool word2vec~\cite{mikolov2013word2vec}.
\item Comparison of 13 state-of-the-art methods on 7 datasets by average precision and processing time. In particular, more correct evaluation of unsupervised methods with many parameters (namely, PU-ATR~\cite{astrakhantsev2015phd} and aforementioned KeyConceptRelatedness) by adapting the cross-validation strategy.
\end{enumerate}

This paper is organized as follows.
Section~\ref{sec:relWork} discuss ATR methods and overview existing software tools for ATR.
Section~\ref{sec:arch} describes ATR4S architecture and implemented methods, including proposed modifications of KeyConceptRelatedness.
Section~\ref{sec:eval} presents experimental evaluation.
The last section outlines the paper.

\section{Related work} \label{sec:relWork}

\subsection{ATR methods}

The first survey by Kageura and Umino~\cite{kageura1996methods} devoted to ATR distinguished all methods into linguistic and statistical.
%aspects of all methods: unithood, connection of words in a term, and termhood, relatedness of term to considered domain. 
Then, in a survey by Pazienza et~al.~\cite{pazienza} of 2005, it was argued that all modern algorithms include linguistic methods as a filtering step. 
Finally, the most recent survey by Astrakhantsev et~al.~\cite{astrakhantsev2015methods} identified the general pipeline of ATR methods: preprocessing, term candidates collection, term candidates scoring, and term candidates ranking.

%then lingustic step
%most modern - 2 setups; the second one is the following pipeline:
%Most modern survey defines 2 possible task definitions in ATR: classification of each occurrence like in NERC and, far more common, extacting set of terms from set of text documents.

Preprocessing transforms input text into a sequence of elements needed for further term candidates extraction; 
most often, each of such elements consists of lemmatized token with attached part of speech tag; 
some works~\cite{judea2014,zhang2016jate} use instead noun phrases obtained by shallow parsing.
%, but it was shown to be not better, while much more computationally expensive.

Term candidates extraction can be seen as a filtering step: it should throw out such words and collocations that are almost certainly not terms, based on simple linguistic and  statistical criteria like presence of stop word or minimal frequency of occurrence. 

Term candidates scoring, i.e. assigning a number to each term candidate reflecting its likelihood of being a term, is the most important and sophisticated step in the whole pipeline.
Considering the type of information used to score term candidate, we can form the following groups:
methods based on frequencies of term candidate occurrences (with large subgroup of word association measures);
on occurrences contexts; 
on reference corpora; 
on topic modeling;
on Wikipedia.
ATR4S includes most promising methods for each group; see details in section~\ref{sec:candScoring}.

Term candidates ranking is the final step, which lets to take the top candidates and thus distinguish terms from not terms.
It is trivial in case of only one term scoring method, because we can simply rank by that score; and the vast majority of works belongs to this group. 
Some works use linear combination, voting algorithm or semi-supervised learning, we discuss them in section~\ref{sec:candRanking}.
Another set of works~\cite{dialog2014,syrcodis2013,lukashevich2013experimental}  apply supervised machine learning.

\subsection{ATR software tools}

%Multiple ATR algorithms have been implemented as software tools to date.
There are many software tools developed for ATR to date.
However, most of them provide only 1 or 2 methods, which are usually outdated.
For example, TerMine\footnote{\url{http://www.nactem.ac.uk/software/
termine}} is based on CValue/NC-Value methods (and academic usage only);
FlexiTerm contains C-Value and "a simple term variant normalisation method"~\cite{spasic2013flexiterm};
TOPIA\footnote{\url{https://pypi.python.org/pypi/topia.termextract}} lists only one method without algorithm description and it is not updated since 2009;
TermRider\footnote{\url{https://gate.ac.uk/projects/neon/termraider.html}} utilizes TF-IDF only;
TermSuite~\cite{cram2016termsuite} ranks candidates by Weirdness method, but focuses on recognizing term variants based on syntactic and morphological patterns.

Some tools are limited by searching for mentions of (named) entities (for example, OpenCalais\footnote{\url{http://www.opencalais.com/about-open-calais/}}) or named entites and Wikipedia concepts (Texterra~\cite{texterra2014}). 
Another tool\footnote{\url{https://bitbucket.org/Meister17/term-extraction}} supports only supervised recognition of 1-word and 2-words terms.

JATE 2.0~\cite{zhang2016jate} is the most similar tool to ATR4S: it is written in Java and also can be natively used in any JVM-based language, contains many ATR algorithms and multiple methods for term candidates collection; it is highly modular and adaptable. 
However, it lacks a lot of actual state-of-the-art methods, namely those based on occurrences contexts, topic models, Wikipedia, and non-trivial ranking algorithms such as Voting~\cite{zhang2008} and PU-ATR~\cite{astrakhantsev2015phd}.
It also depends on Apache Solr\footnote{\url{http://lucene.apache.org/solr/}}, which may simplify its integration to the application that already uses Solr, but may as well complicate its usage as a library.

\section{Architecture}\label{sec:arch}

ATR4S follows general pipeline of term recognition: texts preprocessing, term candidates collection, term candidates scoring, and term candidates ranking. 
Subsections below describe each step in details.

\subsection{Preprocessing}

At the preprocessing step, ATR4s splits input text documents into sentences, tokenizes obtained sentences, and finds part of speech tags and lemmas for obtained tokens. 
%add more details
In order to perform these tasks, ATR4S incorporated 3 external NLP libraries: Stanford CoreNLP~\cite{corenlp},
Emory nlp4j\footnote{\url{https://emorynlp.github.io/nlp4j/}}
and Apache OpenNLP\footnote{\url{http://opennlp.apache.org/}}.
We use the first one in all experiments\footnote{Preliminary experiments show drop of 1-5\% in average precision in case of switching from Stanford CoreNLP, mainly because of part of speech tagging errors; however, note that Stanford CoreNLP is distributed under GPL, while others are licensed under the Apache License, Version 2.0.}.

%other libraries can be used here

\subsection{Term candidates collection}\label{sec:candCollection}

ATR4S extracts consecutive word n-grams of specified orders (by default, from 1 to 4) as term candidates.

Three basic filters can be applied before formation of term candidate occurrence (or \textit{term occurrence}, for brevity):
\begin{enumerate}
\item Noise word filter: keeps term occurrence if all of its lemmas have length not less than the predefined limit and match the predefined regular expression (by default, length limit is 3 characters and regular expression filters out words containing non-alphanumeric characters). This filter is most useful for texts obtained from automatic parsing (e.g. PDF or HTML) and thus containing a lot of noise words. %, which are out-of-vocabulary for part-of-speech tagger and thus tagged as nouns.
\item Stop word filter: keeps term occurrence if the predefined set of stop words contains no lemma of the term occurrence. By default, we use stop words list from the SMART retrieval system~\cite{Salton1971}.
\item Part of speech (PoS) tags pattern: keeps term occurrence if its PoS tags match the pattern encoded as the regular expression. By default, we apply the commonly-used pattern~\cite{Buitelaar2009regexp} extended by allowing prepositions between nouns:
\verb!(NN(S)?|JJ|NNP|NN(S?)IN)*(NN(S)?)! \\
\end{enumerate}

Then ATR4S combines occurrences with the same canonical representation (lemmas joined by underscore symbol, e.g. \verb!information_processing!) as belonging to the same term candidate.

Finally, ATR4S filters out term candidates occurring rarer than the predefined number of times (by default, 2), in order to (a) reduce noise occurring due to errors in preprocessing steps or input data preparation; (b) improve quality of scoring methods: most of them use occurrences frequency; and (c) reduce computation efforts.

%Note that ATR4S does not include noun phrases, because it was shown~\cite{zhang2016jate} to obtain not better quality, while require much more computational efforts.

\subsection{Term candidates scoring}\label{sec:candScoring}

ATR4S includes more than 15 methods for term candidates scoring; %TODO fix constant
below we describe them grouped by the type of information used to score term candidate.

\subsubsection{Methods based on occurrences frequencies}

Most methods for term candidates scoring are based on the intuition that the more frequently some word or collocation occur in the domain-specific text collection, the more likely it is a term for this domain.
This subsection describes methods that utilize this intuition only, i.e. those considering only frequencies of words constituting term candidate and ignoring other information.

Besides Term frequency (TF) itself, this group contains Average term frequency (ATF)~\cite{zhang2016jate}, TF-IDF~\cite{evans1995tfidf},  Residual IDF (RIDF)~\cite{zhang2016jate}, CValue~\cite{frantzi2000automatic}, Basic~\cite{bordea2013domainPaper}, and ComboBasic~\cite{astrakhantsev2015phd}.

ATF simply normalizes term frequency by number of documents containing this term candidate. 
%\begin{equation}\label{eq:atf}
%ATF(t)=\frac{TF(t)}{TF_d(t)},
%\end{equation}
%where $DTF(t)$ is a number of documents containing term candidate $t$.

TF-IDF is a classical information retrieval measure showing high values for term candidates that occur frequently in few documents:
\begin{equation}\label{eq:tfidf}
TF \cdot IDF(t) = TF(t) \cdot \log \frac{D}{DTF(t)},
\end{equation}

where $D$ is a total number of document in collection,
$DTF(t)$ is a number of documents containing term candidate $t$.

RIDF was originally proposed for keywords extraction~\cite{church1999inverse} and than re-used for term recognition~\cite{zhang2016jate}.
It is based on the assumption that the deviation of observed IDF from the IDF modeled by Poisson distribution is higher for keywords than for ordinary words. 

\begin{equation}\label{eq:ridf}
RIDF(t) = TF(t) \cdot \log \frac{D}{DTF(t)} + \log(1-e^{-ATF(t)}),
\end{equation}

C-Value, one of the most popular methods, promotes term candidates that occur frequently, but not as parts of other term candidates.
This method was supposed to work with multi-word term candidates only;
ATR4S includes modification proposed by Ventura et al.~\cite{ventura2013combining} that supports one-word term candidates as well:
 
 \begin{equation}\label{eq:ventura}
C \text{-}Value(t) = 
 \begin{cases}
 \log_{2} (|t|+0.1) \cdot TF(t),&\text{if $\{s : t \subset s\} = \emptyset$;}\\
 \log_{2} (|t|+0.1) \cdot \left( TF(t) - \frac{\sum_{s} TF(s)}{|\{s : t \subset s\}|}\right),&\text{else.}
 \end{cases}
\end{equation}

where $|t|$ is a length of term candidate $t$ (number of words), 
$s$ is a set of term candidates containing $t$, i.e. such candidates that $t$ is a their substring.

Basic is a modification of C-Value for intermediate level (of specificity) terms extraction.
Like original C-Value, it can extract multi-word terms only; 
however, unlike C-Value, Basic promotes term candidates that are part of other term candidates,
because such terms are usually served for creation of more specific terms.

\begin{equation}\label{eq:basic}
Basic(t)=|t|\log f(t) + \alpha e_t,
\end{equation}

where $e_t$ is a number of term candidates containing $t$.

ComboBasic modifies Basic further, so that the level of term specificity can be customized by changing parameters of the method:

\begin{equation}\label{eq:combobasic}
ComboBasic(t)=|t|\log f(t) + \alpha e_t + \beta e'_t,
\end{equation}

where $e'_t$ is a number of term candidates that are contained in $t$.
Therefore, by increasing $\beta$, one can extract more specific terms and vice versa.

Note that ATR4S does not include methods based on word association measures like 

z-test~\cite{dennis1965construction}, t-test~\cite{church19916}, $\chi^2$-test, Loglikelihood~\cite{dunning1993accurate}, Mutual Information ($MI$)~\cite{church1990word}, Lexical Cohesion~\cite{park2002automatic}, Term Cohesion~\cite{kozakov2004glossary},
because they were repeatedly shown to obtain not better results than simple frequency~\cite{wermter2006you,lukashevich2013experimental,zhang2016jate}.

\subsubsection{Methods based on occurrences contexts}

Methods from this group follows the distributional hypothesis~\cite{harris1954distributional}
and try to distinguish terms from non-terms by considering their contexts.
We are aware of only 2 such methods: NC-Value~\cite{frantzi2000automatic} and DomainCoherence~\cite{bordea2013domainPaper}; since the latter is a modification of the former and was shown to work better~\cite{bordea2013domainPaper}, ATR4S includes only it.

DomainCoherence works in 3 steps. 
First, it extracts 200 best term candidates by using Basic method.

Then, words from contexts of previously extracted 200 terms are filtered: it keeps only nouns, adjectives, verbs and adverbs that occur in at least 1 quarter of all documents and are similar to these 200 term candidates, i.e. ranked in the top 50 by averaged Normalized PMI:

\begin{equation}\label{eq:pmi}
s(w) = \frac{1}{|T|}\sum_{t \in T}NPMI(t,w) = \frac{1}{|T|}\sum_{t \in T}\frac{log \left( \frac{P(t,w)}{P(t)P(w)}\right)}{log (P(t,w))},
\end{equation}

where $w$ is a context word; 
$T$ is a set of 200 best term candidates extracted by Basic;
$P(t,w)$ is a probability of occurrence of word $w$ in the context of $t$;
$P(t)$ and $P(w)$ are probabilities of occurrences of term $t$ and word $w$, correspondingly.
These probabilities are estimated on the basis of occurrence frequencies in the input collection; context is considered to be a 5 words window.

Finally, as a weight of a term candidate, DomainCohrerence takes the average of the same NPMI measures computed with each of 50 context words extracted at the previous step.

\subsubsection{Methods based on reference corpora}

There are multiple methods based on the assumption that terms can be distinguished from other words and collocations by comparing occurrences statistics of considered domain-specific collection with statistics of some reference corpus (usually, from general domain).

DomainPertinence~\cite{meijer2014semantic} is the simplest implementation of this idea:
\begin{equation}
DomainPertinence(t) = \frac{TF_{target}(t)}{TF_{reference}(t)},
\end{equation}
where $TF_{target}(t)$ is a frequency of term candidate $t$ in target (domain-specific) collection;
$TF_{reference}$ is a frequency in reference (general) collection.

Weirdness~\cite{ahmad1999university} normalizes it by sizes (in number of words) of document collections:

\begin{equation}
Weirdness(t) = \frac{NTF_{target}(t)}{NTF_{reference}(t)},
\end{equation}
where $NTF_{target}(t)$ and $NTF_{reference}$ are frequencies of $t$ normalized by sizes of target and reference collections, respectively\footnote{Note that in case of simple ranking Weirdness and DomainPertinence show exact the same results, because scores for the same term candidate computed by these 2 methods differ by a constant multiplier only.}.

Relevance~\cite{penas2001corpus} further updates it by taking into account fraction of documents, where term candidate occur:
\begin{equation}\label{eq:relevance}
Relevance(t) = 1 - \left( {\log_2\left(2 + \frac{NTF_{target}(t) \cdot DF_{target}(t)}{NTF_{reference}(t)}\right)}\right) ^{-1}
\end{equation}

ATR4S uses Corpus of Historical American English\footnote{\url{http://www.ngrams.info/download_coha.asp}} as a reference collection. 
%;it contains 1.9 million normalized words and collocations, total size is 1.3 billion.

\subsubsection{Methods based on topic modeling}

These methods are based on the idea that topic modeling uncovers semantic information useful for term recognition; 
in particular, that distribution of words over topics found by topic modeling is a less noisy signal than simple frequency of occurrences.

To the best of our knowledge, this group contains only one method capable to extract terms of arbitrary length, that is Novel Topic Model~\cite{li2013novelTM}.

First, it obtains probability distribution of words over the following topics:
$\phi^t$ -- general topics ($1\leq t \leq 20$);
$\phi^B$ -- background topic;
$\phi^D$ -- document-specific topic.
Then, it extracts 200 words most probable for each topic: $V_t$, $V_B$, $V_D$, correspondingly;
finally, for each term candidate $c_i$ its weight is computed as a sum of maximal probabilities for each of its $L_i$ words ($w_{i1}w_{i2}...w_{iL_i}$):
\begin{equation}\label{eq:novelTM}
NTM(c_i)=\log(TF_i) \cdot \sum_{1\leq j \leq L_i,w_j\in \cup \{V_t\}_{t \in T \cup \{B,D\}}} \phi_{w_j}^{mt_{w_j}},
\end{equation}
where
$mt_{w_j}= \operatorname*{arg\,max}_{t \in T \cup \{B,D\}} \phi_{w_j}^{t}$

For topic modeling, ATR4S uses open source framework\footnote{\url{https://github.com/ispras/tm}}. % for regularized multilingual robust topic modeling.

\subsubsection{Methods based on Wikipedia}

All methods mentioned above require large collection of text documents, 
otherwise statistics of term candidates occurrences is too noisy.
The only way to overcome it in case of small collection is to use external resources.
Such a resource should satisfy two requirements:
(a) it should be specific enough to contain domain-specific information needed to distinguish terms from not terms; 
(b) it should be general enough to be applicable for many domains in practice.
Wikipedia\footnote{\url{https://www.wikipedia.org/}} satisfy these requirements: it is multilingual (English version contains more than 5 million articles), covers a lot of domains and keeps growing.

One of the simplest method in this group is WikiPresence: it returns 1 if term candidate occurs in Wikipedia pages as hyperlink caption; 0 otherwise.
Example of its usage is an additional filter for other methods~\cite{astrakhantsev2015phd}.

LinkProbability~\cite{procOfIspras2014} is a normalized frequency of being hyperlink in Wikipedia pages:

\begin{equation}
LinkProb_T(t)=
 \begin{cases}
 0\text{ - if Wikipedia does not contain $t$ or $\frac{H(t)}{W(t)} < T$;}\\
 \frac{H(t)}{W(t)}\text{ - else,}
 \end{cases}
\end{equation}
where $H(t)$ is a number of occurrences of term candidate $t$ as a hyperlink caption;
$W(t)$ is a total number of occurrences in Wikipedia pages;
$T$ is a method parameter needed to filter out too small values, because they occur due to markup errors in most cases; experimentally chosen value $T=0.018$ is used by default.

This method propagates term candidates that are specific enough to be provided with a hyperlink; however, it is able to distinguish terms from general words and collocations, but not from terms of other domains.

KeyConceptsRelatedness~\cite{procOfIspras2014} interprets domain-specific terms as words and collocations that are semantically related to knowingly domain-specific concepts.
This method assumes concepts that are key for many documents in the input collection to be a good approximation for such knowingly domain-specific concepts.

Originally, it was based on computation of semantic relatedness between two Wikipedia concepts (i.e. pages) by Dice measure, which considers numbers of common and total neighbors of these pages. % ~\cite{procOfIspras2014}.
We modified this: instead of Dice measure, we use cosine distance between word embedding vectors~\cite{mikolov2013word2vec} corresponding to Wikipedia concepts.
More precisely, we preprocess\footnote{\url{https://github.com/phdowling/wiki2vec}} Wikipedia dump by removing markup, while keeping occurrences of Wikipedia concepts (replace each hyperlink by special token that includes title of link's target concept), and by tokenizing and stemming, 
then we build\footnote{\url{https://github.com/RaRe-Technologies/gensim}} word embedding model and, finally, use it for semantic relatedness computation.

Another modification relates to extraction of key concepts from a text document.
Initially, KeyConceptsRelatedness used algorithm based on semantic graph construction and clustering, but it works too slowly: in particular, because it requires full word sense disambiguation of all texts.

We propose to use simplified version of KP-Miner~\cite{el2010kpminer}:
In order to be considered as a candidate to key concept, a word or a collocation must:
(a) occur at least twice; 
(b) have an occurrence among the first 800\footnote{We keep original constant from KP-Miner algorithm.} words;
(c) be a valid term candidate, i.e. satisfy requirements listed in Section~\ref{sec:candCollection};
and (d) be contained in the vocabulary of constructed word embedding model (i.e. Wikipedia dump should contain at least 5 hyperlinks to the concept with the same title as the word/collocation).
Then we rank such candidates by the product of their length (in words) and number of occurrences in the document.

In summary, the modified algorithm for KeyConceptsRelatedness is the following:
\begin{enumerate}
\item Extract key concepts for the whole document collection:
	\begin{enumerate}
	\item extract $d$ key concepts from each document (see the algorithm above);% ($d=3$);
	\item keep $N$ key concepts with maximal frequency (number of being chosen as a key concept).% ($N=200$). 
	\end{enumerate}
\item For each term candidate: if the word embedding model does not contain the term candidate, then return $0$; otherwise compute semantic relatedness to extracted $N$ key concepts by weighted kNN adapted for the case with only positive instances:
\begin{equation}
sim_k(c,C_N)=\frac{1}{k}\sum_{i=1}^k cos(v_c,v_i)
\end{equation}
where $c$ is a term candidate; 
${C_N}$ is a set of $N$ key concepts sorted by semantic relatedness to $c$ in descending order;
$k$ is a parameter from kNN (should be much smaller than $N$);
$v_c$ is an embedding vector corresponding to the term candidate;
$v_i$ is an embedding vector corresponding to the key concept $i$.
\end{enumerate}

This method propagates a term candidate that has corresponding Wikipedia article, which is semantically related to key concepts of the whole document collection.

\subsection{Term candidates ranking}\label{sec:candRanking}

%Term candidates ranking is the final step, which lets to take the top candidates and thus distinguish terms from not terms.
As we already mentioned in section~\ref{sec:relWork}, 
term candidates ranking becomes non-trivial in case of multiple methods for term candidates scoring.
(Following terminology of machine learning, we will refer such methods for scoring as \textit{features}, for brevity.)
General idea for this problem is to aggregate values of multiple features into one number (usually, between 0 and 1), thus reducing the task to ranking by one method.

One of the most popular method is a linear combination of features with some predefined (usually, equal) coefficients. 
Examples include PostRankDC~\cite{bordea2013domainPaper}
%TermExtractor~\cite{sclano2007termextractor},
and GlossEx~\cite{park2002automatic}.

Note that linear combination does not require scores of other term candidates to be computed in advance,
so it is simpler and faster\footnote{In particular, because it can be easily parallelized by term candidates.}, but misses potentially useful information.
Voting algorithm~\cite{zhang2008} considers values of all term candidates and it was shown~\cite{zhang2008} to outperform single methods and weighted average (i.e. linear combination):

\begin{equation}
V(t) = \sum_{i=1}^n \frac{1}{r(f_i(t))},
\end{equation}
where $r(f_i(t))$ is a rank of term candidate $t$ among all candidates sorted by feature $f_i$ only;
$n$ is a total number of aggregated features.

More sophisticated approach is PU-ATR~\cite{procOfIspras2014}, which is based on the ideas of bootstrapping (like NC-Value and DomainCoherence) and positive unlabeled (PU) learning.

It extracts top 50-200 terms by single method (\textit{seed method});
then computes values for multiple features for all term candidates;
learns positive-unlabeled classifier by considering these seed terms as positive instances  and all other term candidates as unlabeled instances, where each instance is a vector of feature values;
and, finally, applies learned classifier to each term candidate, so that the obtained classifier's confidence is a final aggregated value.

ComboBasic is recommended~\cite{astrakhantsev2015phd}
as a seed method, because 
(a) it lets to adjust the level of specificity of seed terms and thus indirectly affects the level of specificity of all terms;
(b) it is simple enough to include terms of different nature, i.e. terms that can be extracted by using different types of information (see Section~\ref{sec:candScoring}), so that PU algorithm overfits less probably.

Note that we perform probabilistic classification, which can suffer from the problem of multicollinearity. Thus, following the previous work~\cite{astrakhantsev2015phd}, we assume that scoring methods from different groups weakly correlate and choose them as features:
C-Value (occurrences frequencies), 
DomainCoherence (occurrences contexts), 
Relevance (reference corpora),
NovelTopicModel (topic modeling),
LinkProbability (Wikipedia, domain-independent specificity),
KeyConceptRelatedness (Wikipedia, domain-specificity). 

% max iterations limit

Different Positive-Unlabeled algorithms were shown~\cite{astrakhantsev2015phd} to work similarly for this task, so we chose the simplest one~\cite{liu2002spy}, with Logistic Regression\footnote{We use Apache Spark MLlib for supervised ML: \url{http://spark.apache.org/mllib/}} as an internal probabilistic classifier (during preliminary experiments we found it outperforming Random Forest classifier).

% (supervised) - not interesting in practice

\subsection{Tool features}

ATR4S is highly scalable (by CPUs of one machine), modular and configurable tool that supports automatic caching.

Scalability is provided by storing documents and candidates in Scala parallel collection: preprocessing and most steps of candidates collection can be parallelized by documents; candidates scoring can be parallelized by candidates themselves.

%Usage of dependency injection pattern (constructor injection) enables configurability: 
The whole pipeline is instantiated by its own configuration class, which contains corresponding configurations for each constituent step, which are configurable in the same way, i.e. by constructor injection, until the final configurations with constants only, not instances of other configuration classes.
This configuration can be serialized/deserialized to/from a human-readable JSON file that can be manually edited, 
so the tool can be easily configured without a necessity to rebuild it.

Described architecture enables automatic caching: since configuration for each step uniquely determines result of such step, we can cache that result and address it by the corresponding configuration\footnote{See details in the source code, class ru.ispras.atr.utils.Cacher.}.
Considering the observation that ATR methods usually require fine-tuning for optimal quality and thus are often launched many times, such caching can significantly speed up unchanged (previous) steps, e.g. dataset preprocessing or candidates collection, and therefore, speed up the whole process.

\section{Evaluation}\label{sec:eval}

\subsection{Experiments design}

We evaluate ATR4S on 7 datasets:
GENIA~\cite{kim2003genia},
FAO~\cite{medelyan2008domain},
Krapivin~\cite{krapivin2009large},
Patents~\cite{judea2014},
ACL~RD-TEC~\cite{acl},
ACL~RD-TEC~2.0~\cite{acl2},
EuroParl~\cite{koehn2005europarl} with Eurovoc thesaurus\footnote{\url{http://eurovoc.europa.eu/drupal}}.
See table~\ref{table:datasets} for summary statistics. 

\begin{table}[h]
\centering
\caption{Datasets summary statistics}
\label{table:datasets}
\begin{tabular}{@{}lcrrrc@{}}
\toprule
\multicolumn{1}{c}{Dataset} & Domain & \multicolumn{1}{c}{Docs} & \multicolumn{1}{c}{\begin{tabular}[c]{@{}c@{}}Words,\\ thousands\end{tabular}} & \multicolumn{1}{c}{\begin{tabular}[c]{@{}c@{}}Expected\\ Terms\end{tabular}} & \begin{tabular}[c]{@{}c@{}}Source\\ of terms\end{tabular} \\ \midrule
GENIA & Biomedicine & 2000 & 494 & 35104 & Manual markup \\
FAO & Agriculture & 779 & 26672 & 1554 & Authors' keywords \\
Krapivin & Computer Science & 2304 & 21189 & 8766 & \begin{tabular}[c]{@{}c@{}}Authors' keywords and\\ Protodog~\cite{faralli2013protodog} glossary\end{tabular} \\
Patents & Engineering & 12 & 120 & 1595 & Manual markup \\
ACL & \begin{tabular}[c]{@{}c@{}}Computational\\ Linguistics\end{tabular} & 10085 & 41202 & 21543 & \begin{tabular}[c]{@{}c@{}}Manual annotation of\\ top 82000 candidates\end{tabular} \\
ACL 2.0 & \begin{tabular}[c]{@{}c@{}}Computational\\ Linguistics\end{tabular} & 300 & 33 & 3095 & Manual markup \\
Europarl & Politics & 9672 & 63279 & 15094 & Eurovoc thesaurus \\ \bottomrule
\end{tabular}
\end{table}

We extract term candidates by using default parameters and filters described in section~\ref{sec:candCollection}; 
table~\ref{table:candidates} shows summary statistics of collected candidates.

\begin{table}[h]
\centering
\caption{Term candidates summary statistics}
\label{table:candidates}
\begin{tabular}{@{}lrrrrrc@{}}
\toprule
\multicolumn{1}{c}{Dataset} & \multicolumn{1}{c}{1-grams} & \multicolumn{1}{c}{2-grams} & \multicolumn{1}{c}{3-grams} & \multicolumn{1}{c}{4-grams} & \multicolumn{1}{c}{\begin{tabular}[c]{@{}c@{}}Total\\ candidates\end{tabular}} & \begin{tabular}[c]{@{}c@{}}Candidates among \\ expected terms\end{tabular} \\ \midrule
GENIA & 5000 & 8536 & 2694 & 506 & 16736 & 9433 \\
FAO & 44835 & 201685 & 52004 & 7925 & 306449 & 1343 \\
Krapivin & 36665 & 153625 & 44488 & 7259 & 242037 & 6038 \\
Patents & 1105 & 1105 & 290 & 47 & 2650 & 729 \\
ACL & 91026 & 236001 & 65195 & 7528 & 399750 & 14903 \\
ACL 2.0 & 763 & 520 & 65 & 6 & 1354 & 755 \\
Europarl & 24040 & 188111 & 40336 & 3292 & 255779 & 6841 \\ \bottomrule
\end{tabular}
\end{table}

We use a standard metric for ATR, that is average precision at level K (AvP): 
\begin{equation}
	AvP(K)=\sum\limits_{i=1}^K P(i)(R(i)-R(i-1)),
\end{equation}
where $P(i)$ is precision at level $i$; 
$R(i)$ is recall at level $i$.

We choose K to be equal to the number of expected terms among extracted term candidates for the dataset (see the last column in table~\ref{table:candidates}):
in this case, perfect algorithm for ranking term candidates reaches 100\% quality by AvP.

In order to find best parameters of the methods modified in this work, 
that are KeyConceptRelatedness and PU-method,
we adapt cross-validation strategy in the following way:
each dataset is considered to be a fold;
one fold is test -- we use it for computing average precision of the parameter set chosen as the best one;
other folds are validation -- we use them for choosing the best parameter set as follows:
\begin{enumerate}
\item evaluate each parameter set on each dataset, i.e. associate each parameter set with the AvP obtained on this dataset;
\item find the maximum AvP on that dataset;
\item compute \textit{relative goodness} of each parameter set for each dataset by dividing AvP of this parameter set on the maximum AvP for this dataset;
\item choose the best (CV) parameter set as the one with maximum product of relative goodnesses over all (validation) datasets.
\end{enumerate}

We prefer this strategy to the commonly used optimization over all datasets, 
because it is more similar to the real setting, when we apply ATR to the new dataset without any labeled data and have to use (or at least start from) parameters that were optimized on some previous datasets.
It is especially relevant for the methods with many parameters like KeyConceptRelatedness and PU-method due to their potentially higher chances of overfitting.
For the same reason, we also keep 1 dataset, Europarl, from using it in any experiments except for the final comparison.

\subsection{Quality results (average precision)}\label{qrAVP}

%\subsubsection{KeyConceptRelatedness}

To optimize parameters of KeyConceptRelatedness method we perform grid search with the following set of possible values:
count of key concepts per document $d = \{3, 5, 10, 15, 20, 30\}$;
total count of key concepts $N = $ \{50, 100, 200, 300, 500\};
count of nearest keys $k = \{1, 2, 3, 5, 10\}$.

\begin{table}[h]
\centering
\caption{Best parameters for KeyConceptRelatedness method}
\label{table:keyConRel}
\begin{tabular}{@{}lrrrrrrrr@{}}
\toprule
\multirow{2}{*}{Test fold} & \multicolumn{3}{c}{Best parameters} & \multicolumn{1}{c}{\multirow{2}{*}{Max AvP}} & \multicolumn{3}{c}{CV parameters} & \multicolumn{1}{l}{\multirow{2}{*}{Test AvP}} \\
 & d & N & k & \multicolumn{1}{c}{} & d & N & k & \multicolumn{1}{l}{} \\ \midrule
GENIA & 3 & 500 & 3 & 0.6845 & 15 & 500 & 2 & 0.6757 \\
FAO & 20 & 300 & 2 & 0.4853 & 15 & 500 & 2 & 0.4671 \\
Krapivin & 15 & 500 & 1 & 0.3623 & 15 & 300 & 2 & 0.2844 \\
Patents & 15 & 50 & 3 & 0.6263 & 15 & 500 & 2 & 0.6190 \\
ACL & 30 & 500 & 2 & 0.3294 & 15 & 500 & 2 & 0.3227 \\
ACL 2.0 & 15 & 200 & 3 & 0.7645 & 15 & 500 & 2 & 0.7124 \\ \bottomrule
\end{tabular}
\end{table}

As we can see from table~\ref{table:keyConRel}, parameters found by cross-validation are quite stable: parameter set $d = 15$, $N=500$, $k=2$ shows the highest result in 4 of 5 cases and in the same 4 cases difference between test AvP and best AvP is about 1-2\%;
at the same time, parameter sets optimized for one dataset (columns named \textit{Best parameters}) predictably vary a lot.
By optimizing over all 6 datasets we have the same parameters set: $d = 15$, $N=500$, $k=2$, 
%which is used for final comparison.
which is used for Europarl dataset in the final comparison methods.

%PU-method

To optimize parameters of PU-ATR method we perform grid search with the following set of possible values:
coefficient used in ComboBasic method for the number of containing terms  $\alpha = \{$0, 0.1, 0.5, 0.75, 1\};
coefficient used in ComboBasic method for the number of contained terms $\beta = \{$0, 0.1, 0.5, -0.1, -0.25, -0.5\};
threshold used in PU algorithm for determining reliable negative instances $t = \{$0.1, 0.05, 0.025\}.

\begin{table}[h]
\centering
\caption{Best parameters for PU-ATR method}
\label{table:puATR}
\begin{tabular}{@{}lrrrrrrrr@{}}
\toprule
\multirow{2}{*}{Test fold} & \multicolumn{3}{c}{Best parameters} & \multicolumn{1}{c}{\multirow{2}{*}{Max AvP}} & \multicolumn{3}{c}{CV parameters} & \multicolumn{1}{l}{\multirow{2}{*}{Test AvP}} \\ \cmidrule(lr){2-4} \cmidrule(lr){6-8}
 & a & b & t & \multicolumn{1}{c}{} & a & b & t & \multicolumn{1}{l}{} \\ \midrule
GENIA & 0.5 & -0.5 & 0.05 & 0.7865 & 0.75 & 0.1 & 0.05 & 0.7823 \\
FAO & 0.75 & 0.1 & 0.1 & 0.4526 & 1.0 & 0.1 & 0.025 & 0.4429 \\
Krapivin & 0.5 & 0.1 & 0.025 & 0.4389 & 0.75 & 0.1 & 0.1 & 0.4210 \\
Patents & 0.75 & -0.5 & 0.025 & 0.6925 & 0.75 & 0.1 & 0.05 & 0.6821 \\
ACL & 0.1 & 0.5 & 0.1 & 0.5089 & 0.75 & 0.1 & 0.05 & 0.4938 \\
ACL 2.0 & 0.5 & 0.5 & 0.025 & 0.8137 & 1.0 & -0.1 & 0.1 & 0.8028 \\ \bottomrule
\end{tabular}
\end{table}

Table~\ref{table:puATR} shows that parameters are not so stable, 
but the difference between test AvP and best AvP is about 1-2\% in all cases. 
By optimizing over all 6 datasets we have the following parameter set: $\alpha = 0.75$, $\beta=0.1$, $t=0.05$, which is used for Europarl dataset in the final comparison methods.

\begin{table}[h]
\centering
\caption{Comparison of all methods over all datasets (by average precision)}
\label{table:finalComp}
\begin{tabular}{@{}lrrrrrrr@{}}
\toprule
Method & GENIA & FAO & Krapivin & Patents & ACL & ACL2 & Europarl \\ \midrule
AvgTermFreq & 0.7105 & 0.0415 & 0.1107 & 0.5397 & 0.0682 & 0.6802 & 0.1689 \\
ResidualIDF & 0.7047 & 0.0133 & 0.1063 & 0.5268 & 0.0645 & 0.6774 & 0.1302 \\
CValue & 0.7283 & 0.3845 & 0.4009 & 0.6452 & 0.4304 & 0.7879 & 0.3213 \\
Basic & 0.6444 & 0.3795 & 0.3912 & 0.5548 & \textbf{0.5393} & 0.6966 & \textbf{0.3917} \\
ComboBasic & 0.6440 & 0.3797 & 0.3913 & 0.5526 & \textbf{0.5391} & 0.7013 & \textbf{0.3920} \\
PostRankDC & 0.6655 & 0.4138 & 0.4068 & 0.5033 & 0.4577 & 0.6471 & 0.3784 \\
Relevance & 0.7410 & 0.1504 & 0.2988 & 0.5044 & 0.4782 & 0.7530 & 0.2139 \\
Weirdness & 0.7672 & 0.1478 & 0.3315 & 0.5422 & 0.4797 & 0.7579 & 0.2270 \\
NovelTopicModel & 0.7138 & 0.0598 & 0.1081 & 0.6003 & 0.2484 & 0.7958 & 0.2076 \\
LinkProbability & 0.7071 & 0.0068 & 0.1024 & 0.4571 & 0.0980 & 0.7185 & 0.0851 \\
KeyConceptRel. & 0.6758 & \textbf{0.4671} & 0.3384 & 0.6190 & 0.3227 & 0.7124 & 0.3408 \\
Voting & 0.7582 & 0.1326 & 0.2683 & 0.6243 & 0.3353 & 0.7871 & 0.2617 \\
PU-ATR & \textbf{0.7823} & 0.4429 & \textbf{0.4210} & \textbf{0.6821} & 0.4938 & \textbf{0.8028} & 0.3688 \\
\bottomrule
\end{tabular}
\end{table}

Table~\ref{table:finalComp} presents comparison of all methods over all datasets.
Note that we use parameter set chosen by cross-validation for KeyConceptRelatedness and PU-ATR and default parameters for other methods. 
Voting aggregates the same 5 features as PU-ATR, see section~\ref{sec:candRanking}.

PU-ATR seems to be the most stable: it is the best for 4 datasets and in top 3 methods for all datasets. 
However, it is the most computationally intensive method, see the next subsection. % with not so high difference)
%One of the reasons that Basic is the best for 2 datasets may be the fact that this is the only method ignoring one-word terms.

Note also that none of the methods showing best results in this experiment are implemented in other tools.

\subsection{Performance results (time)}

We estimated performance of ATR4S on a machine with Intel Core i5-2500 (3.3GHz, 4 cores) and 32 Gb RAM, from which 12 Gb was set as a maximum memory allocation pool for Java Virtual Machine, see Table~\ref{table:time}.
Since preprocessing and candidates collection steps are the same for all methods,
we show them in the first 2 rows and ignore that time for scoring/ranking methods.

\begin{table}[h]
\centering
\caption{Comparison of all methods over all datasets (by time, in \textbf{s}econds or \textbf{m}inutes)}
\label{table:time}
\begin{tabular}{@{}lrrrrrrr@{}}
\toprule
Task & GENIA & FAO & Krapivin & Patents & ACL & ACL2 & Europarl \\ \midrule
Preprocessing & 6.6s & 7.1m & 4.6m & 3.7s & 7.8m & 2.0s & 8.1m \\
Candidates & 5.3s & 3.4m & 2.2m & 1.7s & 4.5m & 1.1s & 6.9m \\ \cmidrule(r){1-1}
AvgTermFreq & 0.1s & 0.6s & 0.5s & 0.0s & 0.8s & 0.0s & 0.8s \\
ResidualIDF & 0.1s & 0.6s & 0.5s & 0.0s & 0.8s & 0.0s & 0.8s \\
CValue & 0.2s & 1.8s & 1.6s & 0.1s & 2.3s & 0.1s & 1.5s \\
Basic & 0.1s & 1.0s & 0.8s & 0.1s & 1.2s & 0.1s & 0.8s \\
ComboBasic & 0.2s & 1.9s & 1.5s & 0.1s & 2.2s & 0.1s & 1.6s \\
PostRankDC & 1.1s & 35.9s & 12.9s & 0.6s & 21.2s & 0.4s & 18.2s \\
Relevance & 1.4s & 2.1s & 2.1s & 1.8s & 2.4s & 2.2s & 2.3s \\
Weirdness & 1.6s & 1.8s & 1.8s & 1.6s & 1.8s & 2.3s & 1.8s \\
NovelTopicModel & 32.5s & 8.7m & 8.0m & 2.3s & 23.6m & 3.4s & 21.3m \\
LinkProbability & 15.5s & 20.3s & 17.6s & 18.9s & 18.6s & 16.0s & 20.1s \\
KeyConceptRel. & 1.1m & 5.6m & 3.5m & 1.1m & 6.3m & 1.0m & 9.8m \\
Voting & 2.5m & 15.1m & 13.1m & 2.0m & 32.7m & 1.9m & 30.5m \\
PU-ATR & 2.5m & 16.5m & 12.4m & 1.9m & 34.0m & 1.9m & 33.3m \\ \bottomrule
\end{tabular}
\end{table}

As we can see, methods from the first 3 groups, i.e. those based on occurrences frequencies, contexts and reference corpora, are the fastest.
Methods based on Wikipedia require constant 15 sec (LinkProbability) or 1 min (KeyConceptRelatedness) for initialization, then their times depend on dataset size almost linearly.
NovelTopicModel is the slowest for big datasets; however, its average precision is not good for big datasets anyway.
Time required for PU-ATR is almost the sum of used features times and Spark start time (about 30 secs).

%TODO comparison with JATE 2.0 on the same features on genia
%time and (opt) avp

\section{Conclusion}\label{sec:conclusion}

This paper presents ATR4S, an open-source tool for automatic terms recognition, and experimental comparison of 13 state-of-the-art methods for ATR on 7 datasets.
ATR4S comprises more than 15 methods for ATR, 
supports caching and human-readable configuration; 
it is written in Scala with parallel collections wherever appropriate, so it utilizes all CPU cores.

Experimental comparison confirms observations that (1) no single method is best for all datasets~\cite{zhang2008} and (2) multiple features should be combined for better quality~\cite{syrcodis2013}; 
also it shows that other available tools lack the best methods, i.e. actual state-of-the-art methods, namely 
%PU-ATR, KeyConceptRelatedness, NovelTopicModel, and Basic.
PU-ATR~\cite{astrakhantsev2015phd}, KeyConceptRelatedness~\cite{astrakhantsev2015phd}, NovelTopicModel~\cite{li2013novelTM}, and Basic~\cite{bordea2013domainPaper}.

%created tool (list all features)
It is obvious that ATR4S does not include all methods capable to outperform already implemented ones on some settings, 
but we believe that these implementations can be used as a basis for development of other methods or, at least, for easy comparison.
Nevertheless, the addition of new methods and their experimental evaluation are the main directions of further improvement.
%thus we consider the addition of new methods and their experimental evaluation to be the main directions of further improvement.

Regarding practical aspects of ATR task, 
in particular noisy input datasets, which often contain documents from multiple domains, and scenarios assuming terminology enrichment instead of extraction, 
we believe that incorporation of document clustering and more sophisticated semi-supervised methods are among the most promising research topics.

\begin{acknowledgements}
%If you'd like to thank anyone, place your comments here
%and remove the percent signs.
The author would like to thank Yaroslav Nedumov and Denis Turdakov for their valuable comments and Denis Fedorenko for his help in implementing the previous versions of ATR tool.
\end{acknowledgements}

% BibTeX users please use one of
%\bibliographystyle{spbasic}      % basic style, author-year citations
\bibliographystyle{spmpsci}      % mathematics and physical sciences
\bibliography{atr4s}   % name your BibTeX data base

\begin{thebibliography}{10}
\providecommand{\url}[1]{{#1}}
\providecommand{\urlprefix}{URL }
\expandafter\ifx\csname urlstyle\endcsname\relax
  \providecommand{\doi}[1]{DOI~\discretionary{}{}{}#1}\else
  \providecommand{\doi}{DOI~\discretionary{}{}{}\begingroup
  \urlstyle{rm}\Url}\fi

\bibitem{ahmad1999university}
Ahmad, K., Gillam, L., Tostevin, L., et~al.: University of surrey participation
  in trec8: Weirdness indexing for logical document extrapolation and retrieval
  (wilder).
\newblock In: The Eighth Text REtrieval Conference (TREC-8) (1999)

\bibitem{procOfIspras2014}
Astrakhantsev, N.: Automatic term acquisition from domain-specific text
  collection by using wikipedia.
\newblock Proceedings of the Institute for System Programming \textbf{26}(4),
  7--20 (2014)

\bibitem{astrakhantsev2015phd}
Astrakhantsev, N.: Methods and software for terminology extraction from
  domain-specific text collection.
\newblock Ph.D. thesis, Institute for System Programming of Russian Academy of
  Sciences (2015)

\bibitem{dialog2014}
Astrakhantsev, N., Fedorenko, D., Turdakov, D.: Automatic enrichment of
  informal ontology by analyzing a domain-specific text collection.
\newblock Computational Linguistics and Intellectual Technologies: Papers from
  the Annual International Conference “Dialogue” \textbf{13}, 29--42 (2014)

\bibitem{astrakhantsev2015methods}
Astrakhantsev, N., Fedorenko, D., Turdakov, D.Y.: Methods for automatic term
  recognition in domain-specific text collections: A survey.
\newblock Programming and Computer Software \textbf{41}(6), 336--349 (2015)

\bibitem{bordea2013domainPaper}
Bordea, G., Buitelaar, P., Polajnar, T.: Domain-independent term extraction
  through domain modelling.
\newblock In: the 10th International Conference on Terminology and Artificial
  Intelligence (TIA 2013), Paris, France (2013)

\bibitem{Buitelaar2009regexp}
Buitelaar, P., Eigner, T.: Expertise mining from scientific literature.
\newblock In: Proceedings of the Fifth International Conference on Knowledge
  Capture, K-CAP '09, pp. 171--172. ACM, New York, NY, USA (2009)

\bibitem{church1999inverse}
Church, K., Gale, W.: Inverse document frequency (idf): A measure of deviations
  from poisson.
\newblock In: Natural language processing using very large corpora, pp.
  283--295. Springer (1999)

\bibitem{church19916}
Church, K., Gale, W., Hanks, P., Kindle, D.: 6. using statistics in lexical
  analysis.
\newblock Lexical acquisition: exploiting on-line resources to build a lexicon
  p. 115 (1991)

\bibitem{church1990word}
Church, K.W., Hanks, P.: Word association norms, mutual information, and
  lexicography.
\newblock Computational linguistics \textbf{16}(1), 22--29 (1990)

\bibitem{cram2016termsuite}
Cram, D., Daille, B.: Termsuite: Terminology extraction with term variant
  detection.
\newblock ACL 2016 p.~13 (2016)

\bibitem{dennis1965construction}
Dennis, S.F.: The construction of a thesaurus automatically from a sample of
  text.
\newblock In: Proceedings of the Symposium on Statistical Association Methods
  For Mechanized Documentation, Washington, DC, pp. 61--148 (1965)

\bibitem{dunning1993accurate}
Dunning, T.: Accurate methods for the statistics of surprise and coincidence.
\newblock Computational linguistics \textbf{19}(1), 61--74 (1993)

\bibitem{el2010kpminer}
El-Beltagy, S.R., Rafea, A.: Kp-miner: Participation in semeval-2.
\newblock In: Proceedings of the 5th international workshop on semantic
  evaluation, pp. 190--193. Association for Computational Linguistics (2010)

\bibitem{evans1995tfidf}
Evans, D.A., Lefferts, R.G.: Clarit-trec experiments.
\newblock Information processing \& management \textbf{31}(3), 385--395 (1995)

\bibitem{faralli2013protodog}
Faralli, S., Navigli, R.: Growing multi-domain glossaries from a few seeds
  using probabilistic topic models.
\newblock In: EMNLP, pp. 170--181 (2013)

\bibitem{syrcodis2013}
Fedorenko, D., Astrakhantsev, N., Turdakov, D.: Automatic recognition of
  domain-specific terms: an experimental evaluation.
\newblock In: Proceedings of SYRCoDIS 2013, pp. 15--23 (2013)

\bibitem{frantzi2000automatic}
Frantzi, K., Ananiadou, S., Mima, H.: Automatic recognition of multi-word
  terms:. the c-value/nc-value method.
\newblock International Journal on Digital Libraries \textbf{3}(2), 115--130
  (2000)

\bibitem{gaussier1998flow}
Gaussier, {\'E}.: Flow network models for word alignment and terminology
  extraction from bilingual corpora.
\newblock In: Proceedings of the 17th international conference on Computational
  linguistics-Volume 1, pp. 444--450. Association for Computational Linguistics
  (1998)

\bibitem{harris1954distributional}
Harris, Z.S.: Distributional structure.
\newblock Word \textbf{10}(2-3), 146--162 (1954)

\bibitem{judea2014}
Judea, A., Sch\"{u}tze, H., Bruegmann, S.: Unsupervised training set generation
  for automatic acquisition of technical terminology in patents.
\newblock In: Proceedings of COLING 2014, the 25th International Conference on
  Computational Linguistics: Technical Papers, pp. 290--300. Dublin City
  University and Association for Computational Linguistics, Dublin, Ireland
  (2014)

\bibitem{kageura1996methods}
Kageura, K., Umino, B.: Methods of automatic term recognition: A review.
\newblock Terminology \textbf{3}(2), 259--289 (1996)

\bibitem{kim2003genia}
Kim, J.D., Ohta, T., Tateisi, Y., Tsujii, J.: Genia corpus--a semantically
  annotated corpus for bio-textmining.
\newblock Bioinformatics \textbf{19}(Suppl 1), i180--i182 (2003).
\newblock \doi{10.1093/bioinformatics/btg1023}

\bibitem{koehn2005europarl}
Koehn, P.: Europarl: A parallel corpus for statistical machine translation.
\newblock In: MT summit, vol.~5, pp. 79--86 (2005)

\bibitem{kozakov2004glossary}
Kozakov, L., Park, Y., Fin, T., Drissi, Y., Doganata, Y., Cofino, T.: Glossary
  extraction and utilization in the information search and delivery system for
  ibm technical support.
\newblock IBM Systems Journal \textbf{43}(3), 546--563 (2004)

\bibitem{krapivin2009large}
Krapivin, M., Autaeu, A., Marchese, M.: Large dataset for keyphrases extraction
   (2009).
\newblock
  \urlprefix\url{http://eprints.biblio.unitn.it/1671/1/disi09055-krapivin-autayeu-marchese.pdf}

\bibitem{li2013novelTM}
Li, S., Li, J., Song, T., Li, W., Chang, B.: A novel topic model for automatic
  term extraction.
\newblock In: Proceedings of the 36th international ACM SIGIR conference on
  Research and development in information retrieval, pp. 885--888. ACM (2013)

\bibitem{lingpeng2005improving}
Lingpeng, Y., Donghong, J., Guodong, Z., Yu, N.: Improving retrieval
  effectiveness by using key terms in top retrieved documents.
\newblock In: Advances in Information Retrieval, pp. 169--184. Springer (2005)

\bibitem{liu2002spy}
Liu, B., Lee, W.S., Yu, P.S., Li, X.: Partially supervised classification of
  text documents.
\newblock In: ICML, vol.~2, pp. 387--394. Citeseer (2002)

\bibitem{corenlp}
Manning, C.D., Surdeanu, M., Bauer, J., Finkel, J., Bethard, S.J., McClosky,
  D.: The {Stanford} {CoreNLP} natural language processing toolkit.
\newblock In: Association for Computational Linguistics (ACL) System
  Demonstrations, pp. 55--60 (2014)

\bibitem{mayorov2015aspect}
Mayorov, V., Andrianov, I., Astrakhantsev, N., Avanesov, V., Kozlov, I.,
  Turdakov, D.: A high precision method for aspect extraction in russian.
\newblock In: Proceedings of International Conference Dialog, vol.~2 (2015)

\bibitem{medelyan2008domain}
Medelyan, O., Witten, I.H.: Domain-independent automatic keyphrase indexing
  with small training sets.
\newblock Journal of the American Society for Information Science and
  Technology \textbf{59}(7), 1026--1040 (2008)

\bibitem{meijer2014semantic}
Meijer, K., Frasincar, F., Hogenboom, F.: A semantic approach for extracting
  domain taxonomies from text.
\newblock Decision Support Systems \textbf{62}, 78--93 (2014)

\bibitem{mikolov2013word2vec}
Mikolov, T., Sutskever, I., Chen, K., Corrado, G.S., Dean, J.: Distributed
  representations of words and phrases and their compositionality.
\newblock In: Advances in neural information processing systems, pp. 3111--3119
  (2013)

\bibitem{lukashevich2013experimental}
Nokel, M., Loukachevitch, N.: An experimental study of term extraction for real
  information-retrieval thesauri.
\newblock In: Proceedings of 10th International Conference on Terminology and
  Artificial Intelligence, pp. 69--76 (2013)

\bibitem{park2002automatic}
Park, Y., Byrd, R., Boguraev, B.: Automatic glossary extraction: beyond
  terminology identification.
\newblock In: Proceedings of the 19th international conference on Computational
  linguistics-Volume 1, pp. 1--7. Association for Computational Linguistics
  (2002)

\bibitem{pazienza}
Pazienza, M., Pennacchiotti, M., Zanzotto, F.: Terminology extraction: an
  analysis of linguistic and statistical approaches.
\newblock Knowledge Mining pp. 255--279 (2005)

\bibitem{penas2001corpus}
Pe{\~n}as, A., Verdejo, F., Gonzalo, J., et~al.: Corpus-based terminology
  extraction applied to information access.
\newblock In: Proceedings of Corpus Linguistics, vol. 2001. Citeseer (2001)

\bibitem{acl2}
{QasemiZadeh}, B., Schumann, A.K.: The {ACL} {RD}-{TEC} 2.0 (2016).
\newblock \urlprefix\url{http://hdl.handle.net/11372/LRT-1661}.
\newblock {LINDAT}/{CLARIN} digital library at the Institute of Formal and
  Applied Linguistics, Charles University in Prague

\bibitem{Salton1971}
Salton, G.: The SMART Retrieval System---Experiments in Automatic Document
  Processing.
\newblock Prentice-Hall, Inc., Upper Saddle River, NJ, USA (1971)

\bibitem{spasic2013flexiterm}
Spasi{\'c}, I., Greenwood, M., Preece, A., Francis, N., Elwyn, G.: Flexiterm: a
  flexible term recognition method.
\newblock Journal of biomedical semantics \textbf{4}(1), 1 (2013)

\bibitem{texterra2014}
Turdakov, D.Y., Astrakhantsev, N., Nedumov, Y.R., Sysoev, A., Andrianov, I.,
  Mayorov, V., Fedorenko, D., Korshunov, A., Kuznetsov, S.D.: Texterra: A
  framework for text analysis.
\newblock Programming and Computer Software \textbf{40}(5), 288--295 (2014)

\bibitem{ventura2013combining}
Ventura, J.A.L., Jonquet, C., Roche, M., Teisseire, M., et~al.: Combining
  c-value and keyword extraction methods for biomedical terms extraction.
\newblock In: LBM'2013: International Symposium on Languages in Biology and
  Medicine, pp. 45--49 (2013)

\bibitem{wermter2006you}
Wermter, J., Hahn, U.: You can't beat frequency (unless you use linguistic
  knowledge): a qualitative evaluation of association measures for collocation
  and term extraction.
\newblock In: Proceedings of the 21st International Conference on Computational
  Linguistics and the 44th annual meeting of the Association for Computational
  Linguistics, pp. 785--792. Association for Computational Linguistics (2006)

\bibitem{acl}
Zadeh, B.Q., Handschuh, S.: The acl rd-tec: A dataset for benchmarking
  terminology extraction and classification in computational linguistics.
\newblock In: COLING 2014: Proceedings of the 4th International Workshop on
  Computational Terminology (CompuTerm'14). Dublin, Ireland (2014)

\bibitem{zhang2008}
Zhang, Z., Brewster, C., Ciravegna, F.: A comparative evaluation of term
  recognition algorithms.
\newblock In: Proceedings of the Sixth International Conference on Language
  Resources and Evaluation (LREC08), Marrakech, Morocco (2008)

\bibitem{zhang2016jate}
Zhang, Z., Gao, J., Ciravegna, F.: Jate 2.0: Java automatic term extraction
  with apache solr.
\newblock The LREC 2016 Proceedings  (2016)

\end{thebibliography}

\end{document}